%% file: neurips_2023.tex
\documentclass{article}

    \PassOptionsToPackage{numbers, square}{natbib}

    \usepackage[preprint]{neurips_2023}

\usepackage[utf8]{inputenc} 
\usepackage[T1]{fontenc}    
\usepackage{hyperref}       
\usepackage{url}            
\usepackage{booktabs}       
\usepackage{amsfonts}       
\usepackage{nicefrac}       
\usepackage{microtype}      
\usepackage{xcolor}         
\usepackage{graphicx}
\usepackage{multirow}
\usepackage{caption}
\title{Explaining knock-on effects of bias mitigation}

\author{%
  Svetoslav Nizhnichenkov \\
  IBM Research \\
  School of Computer Science \\
  University College Dublin \\
  Dublin, Ireland \\
  \texttt{svetoslav.nizhnichenkov@ibm.com} \\
  \And
  Rahul Nair \\
  IBM Research\\
  Dublin, Ireland \\
  \texttt{rahul.nair@ie.ibm.com} \\
  \AND
  Elizabeth Daly \\
  IBM Research \\
  Dublin, Ireland \\
  \texttt{elizabeth.daly@ie.ibm.com} \\
  \And
  Brian Mac Namee \\
  School of Computer Science \\
  University College Dublin \\
  Dublin, Ireland \\
  \texttt{brian.macnamee@ucd.ie} \\
}

\begin{document}

\maketitle

\begin{abstract}

In machine learning systems, bias mitigation approaches aim to make outcomes fairer across privileged and unprivileged groups. Bias mitigation methods work in different ways and have known ``waterfall'' effects, e.g., mitigating bias at one place may manifest bias elsewhere. In this paper, we aim to characterise impacted cohorts when mitigation interventions are applied. To do so, we treat intervention effects as a classification task and learn an explainable meta-classifier to identify cohorts that have altered outcomes. We examine a range of bias mitigation strategies that work at various stages of the model life cycle. We empirically demonstrate that our meta-classifier is able to uncover impacted cohorts. Further, we show that all tested mitigation strategies negatively impact a non-trivial fraction of cases, i.e., people who receive unfavourable outcomes solely on account of mitigation efforts. This is despite improvement in fairness metrics. We use these results as a basis to argue for more careful audits of static mitigation interventions that go beyond aggregate metrics. 
    
\end{abstract}

\section{Introduction}
\label{introduction}

In the context of decision-making, \textbf{fairness} \cite{mehrabiSurveyBiasFairness2022}  is the absence of any prejudice or favouritism toward an individual or a group based on their inherent or acquired characteristics, while \textbf{bias} \cite{mehrabiSurveyBiasFairness2022} occurs when an algorithm skews its decisions toward a particular group of individuals. Different types of biases can manifest themselves in many shapes and forms.

\emph{Biases in data:}    
Data plays a significant role in the functionality of a wide range of AI systems. If data is biased in any way, the underlying algorithms within these systems will learn these biases and the generated predictions will reflect them. In automation settings, this can lead to biases being perpetuated at scale, along with the inequities they cause.

\emph{Biases in algorithms:} 
Algorithms used to train AI systems can behave in a biased fashion. Often, this is due to specific design choices. For instance, a model trained to predict healthcare costs rather than illness will be biased against communities with low resources.
The outcome of such systems may be used as input into real-life systems and affect the decisions of the people using them. This will also inevitably result in more data with underlying biases that will be used for training different algorithms further down the pipeline.

\emph{Biases in user experience:} Users can promote biased behaviour when interacting with systems. For example, users of a search engine will likely interact with the results closer to the top of a list of results. Should this engagement signal be used as a proxy for popularity and the underlying system retrained to include this information, the results will exhibit a popularity bias where top results become even more popular \cite{lermanLeveragingPositionBias2014}.

To address these different biases, different algorithms have been proposed in the literature along with several notions of (statistical) fairness and associated metrics \cite{catonFairnessMachineLearning2020}. A fairness metric aims to assess how biased a model is and specific mitigation methods improve these metrics by altering data, algorithms, or outcomes. From the perspective of a machine learning practice, mitigation methods can be viewed as being able to work at three different stages of the machine learning pipeline: (1) pre-processing (on the data), (2) in-processing (during the model training phase), or (3) post-processing (adjusting the outcomes). 

However, most mitigation methods are known to have a ``waterfall'' effect, i.e., bias addressed at one place can manifest biases elsewhere in the data \cite{krcoWhenMitigatingBias2023}. Since mitigations are generally viewed as once-off static interventions, these knock-on waterfall effects have rarely been investigated beyond aggregate fairness measures. This paper aims to provide a characterisation of mitigation interventions in terms of impacted cohorts.

The paper makes two contributions. First, we develop a method that uses a supervised meta-classifier to describe impacted cohorts after bias mitigation. The method provides interpretable summaries of cohorts in the form of conjunctions. Second, we demonstrate over a wide range of mitigation strategies, fairness metrics and several datasets, that the meta-classifier is discriminative and highlight several important findings that call for a more careful audit of static mitigation interventions. Specifically, we empirically, show that there is always a negatively impacted cohort of individuals that are impacted solely on account of bias mitigation efforts, and some methods work consistently better than others.

The rest of this paper proceeds as follows. Section \ref{background} introduces background information and related work. Section \ref{methodology} describes the methodological approaches used in our experiments. Section \ref{data} describes the characteristics of the datasets used in our experiments.  Section \ref{results} presents and discusses the results of our experiments. Section \ref{conclusion} summarises the contributions of the paper.

\section{Background \& Related Work}
\label{background}

Our analysis is based on several mitigation methods from the literature. We use the AI Fairness 360 toolkit \footnote{\label{aif360}https://github.com/Trusted-AI/AIF360/} implementations for these methods. 

\emph{Pre-processing} bias-mitigation algorithms work on the raw data by altering it in some way 
    in order to reduce or eliminate the bias present in the data before this data is used for the
    training of some model. The pre-processing fairness-intervention techniques used in this study are Learning Fair Representations (LFR) \cite{zemelLearningFairRepresentations} and Disparate Impact Remover (DIR) \cite{feldmanCertifyingRemovingDisparate2015a}.
    
\emph{In-processing} bias-mitigation algorithms have two parallel goals: accuracy and fairness. This means that the model should display sufficient accuracy while also being fair w.r.t. the fairness constraints. The idea is that such models directly take fairness into account and produce a classifier that will yield less bias compared to a model that is unaware of fairness. The in-processing approaches adopted in this study are Gerry Fair Classification (GF) \cite{kearnsEmpiricalStudyRich2019} \cite{kearnsPreventingFairnessGerrymandering} and Prejudice Remover (PR) \cite{kamishimaFairnessAwareClassifierPrejudice2012a}.
    
\emph{Post-processing} bias-mitigation algorithms work on the predictions already made by a 
    biased model and alter them in some way (e.g., change the target label of an instance from
    positive to negative or vice-versa) to reduce bias according to some fairness metric. The post-processing bias-mitigation approaches utilised in this study are Reject Option Classification (ROC) \cite{kamiranDecisionTheoryDiscriminationAware2012}, Equalized Odds (EO) \cite{hardtEqualityOpportunitySupervised} and Calibrated Equalized Odds (CEO) \cite{pleissFairnessCalibration}.


Several other works have studied similar problems. In \cite{krcoWhenMitigatingBias2023}, the authors compared different fairness strategies to investigate their behaviour at the prediction level w.r.t. whether similar-performing techniques mitigate bias in the same way, impact a similar volume of people and if the same individuals are being affected. The results show that the fairness approaches do in fact impact a different number of individuals as well as even having different targeted cohorts. Furthermore, these observations hold true for multiple executions of the same fairness approaches. 

Friedler et al. \cite{friedlerComparativeStudyFairnessenhancing2019} compared various fairness approaches to uncover the differences w.r.t. performance, and whether fairness interventions have knock-on impacts. The findings indicate that, due to the fact that some portion of the fairness metrics correlate with one another, when a fairness approach optimizes one metric, it also performs well on all other correlated metrics. Moreover, the study shows that the fairness approaches have a tendency to be sensitive to variations in input (e.g., different train/test splits) which results in fluctuating fairness metrics for different splits. Finally, the outcomes of the different fairness approaches were observed to vary a lot even if a given fairness metric is being satisfied due to the underlying difference in the mechanisms of the fairness approaches. 

Marchiori et al. \cite{marchiorimanerbaInvestigatingDebiasingEffects2022} utilised an evaluation framework to perform a comparative study on the analysis of the effect of classification and explainability of two pre-processing bias-mitigation strategies, namely preferential sampling \cite{kamiranClassificationNoDiscrimination} and uniform sampling \cite{kamiranDataPreprocessingTechniques2012a} on a set of known biased datasets. They provided a two-fold quantitative assessment, global high-level assessment and local-based explanations via SHAP \cite{lundbergUnifiedApproachInterpreting} and LIME \cite{ribeiroWhyShouldTrust2016}. Results show the most variation w.r.t. the different classifier was displayed by the fairness metric output while the preferential sampling strategy yielded better results than uniform sampling at reducing the importance of the attributes with sensitive nature.

\cite{aasheimBiasMitigationAIF360} conducted an empirical analysis on a binary classification task dataset \cite{us_census_dataset} to compare the results of two bias-mitigation techniques, ROC and PR, w.r.t. group fairness and accuracy metrics while exploring different protected attribute settings (e.g., \textit{gender}, \textit{race}). Results showed that, overall, ROC outperformed PR based on the metric results. However, PR was found to improve fairness metrics better when changes to its hyperparameter settings were made, but this came at the cost of decreased accuracy. While this study looked at how bias-mitigation approaches compare when only aggregate metrics were taken into account, our study aims to dig deeper and uncover unintended consequences of bias-mitigation strategies that go beyond accuracy and fairness metrics.

\section{Methodology}
\label{methodology}

We focus on binary classification tasks with sensitive attributes 
that divide a population into privileged and unprivileged groups.
The classification assigns each instance in a given group a favourable or an unfavourable label. 
While there are different notions of fairness, e.g., individual fairness (treating similar individuals the same way) and group fairness, we focus on the latter where the objective is to optimize a metric amongst the divided groups. We choose to optimize four different fairness metrics when it comes to group fairness, namely disparate impact, average odds, equal opportunity difference, and statistical parity difference. Finally, we use a batter of mitigation strategies that work at different stages of the model life cycle.

\subsection{Bias-Mitigation Pipeline}

The bias-mitigation pipeline is a mapping $g: D_{x} \rightarrow D_{z}$ that maps the input dataset, $D_{x}$, containing ground truth labels $y$ and predictions $y'$ generated from a model $f$, to $D_{z}$, an output dataset where predictions $y'$ have been changed to bias-mitigated predictions $y''$ as a result of applying some bias-mitigation model $g$.

\begin{figure}[ht]
\centering
\includegraphics[width=0.9\textwidth]{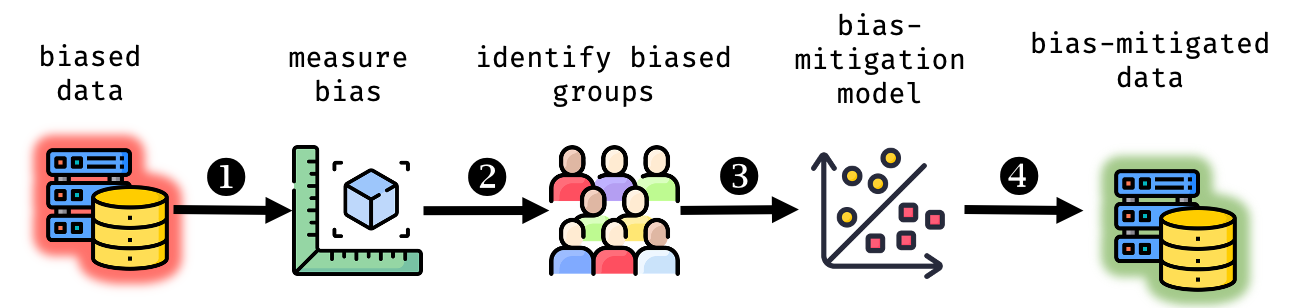}
\caption{Bias-mitigation Pipeline. This pipeline depicts how a dataset with ground truth and predictions from a biased model is utilised to produce a new set of bias-mitigated predictions.}
\label{fig:bias-mit-pipeline}
\end{figure}

Figure \ref{fig:bias-mit-pipeline} provides a graphical overview. Given the dataset $D_{x}$, a fairness metric is used to measure the amount of bias present in it relative to a known privileged group. Following that, Multi-Dimensional Subset Scan \cite{mdss} is performed to identify the cohorts in the data that show the most significant amount of bias. The resulting information regarding the biased cohorts is then passed to the bias-mitigation model, $g$, along with the dataset, $D_{x}$, and it produces a new set of bias-mitigated predictions, $y''$, which replace the predictions, $y'$, that came with the original dataset, thus outputting a dataset, $D_{z}$, containing bias-mitigated predictions, and a model, $g$, which can be used to debias future predictions.

\subsection{Meta-classifier Pipeline}

This pipeline is a mapping $D_{z} \rightarrow h$ where $D_{z}$ is the output dataset from the bias-mitigation pipeline that contains bias-mitigated predictions $y''$ and $h$ is a meta classifier (in this case a decision tree) that has been trained on a mapping between the ground truth and the bias-mitigated predictions. 

\begin{figure}[ht]
\centering
\includegraphics[width=0.7\textwidth]{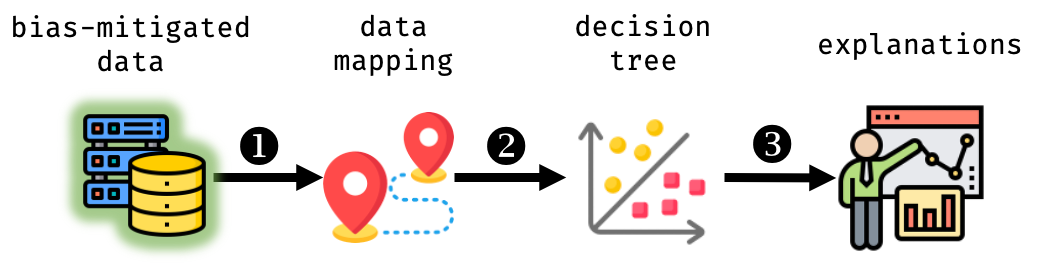}
\caption{Meta-classifier Pipeline. This pipeline displays how an explainable model is used to learn a mapping of bias-mitigated predictions in order to generate explanations that can be used to infer the decision-making process of the bias-mitigation model.}
\label{fig:meta-clf-pipeline}
\end{figure}

Figure \ref{fig:meta-clf-pipeline} gives a high-level graphical overview of the meta-classifier pipeline, the aim of which is to learn a meta classifier in the form of an explainable decision tree, $h$, on a mapping between the ground truth and the bias-mitigated data, and generate explanations to reveal insights about the decision-making of the bias-mitigation model $g$. Given the dataset $D_{z}$, this pipeline applies a mapping between the ground truth, $y$, and the bias-mitigated predictions, $y''$, to produce a new output set, $y'''$, capturing observational changes in treatment. For instance, all individuals whose treatment didn't differ from the ground truth after applying bias mitigation will be assigned the value of 0, whereas the individuals whose treatment changed for the positive (e.g., an individual from the Adult dataset being predicted to earn more than 50,000 USD per annum post bias mitigation while their ground truth states otherwise) will be assigned the value of 1, and the people whose treatment changed for the negative (e.g., an individual from the Utrecht dataset being rejected for the job they applied post bias mitigation while their ground truth states they were approved) will be assigned the value of -1. As a result, this newly obtained set would capture changes in treatment between the ground truth and the bias-mitigated predictions.

Thus, this new output set, $y'''$, will be imposed as the new ground truth for the decision tree to learn from in order to capture the decision-making process of the bias-mitigation model, $g$, and be able to produce explanations in the form of decision trees or boolean decision rules that can be used to identify cohorts which will be affected in different ways (e.g., positively, negatively or no change in treatment).

\section{Data}
\label{data}

There are three different datasets utilised in this study. Most of them are well-known in the fairness research literature. The task associated with each of the datasets is binary classification and they all contain sensitive attributes of some nature (e.g., \textit{age}, \textit{race}, \textit{gender} and \textit{nationality}). The datasets used are: 

\emph{Utrecht Fairness Recruitment:} The Utrecht Fairness Recruitment dataset \cite{utrecht_dataset} is a synthetic dataset containing information about 4,000 individuals. Each individual is described by a number of sensitive attributes (e.g., \textit{gender}, \textit{age}, \textit{nationality}) and general attributes (e.g., \textit{highest degree achieved,} \textit{degree score}, \textit{programming experience}, \textit{sport}, \textit{international experience}). The target specifies whether that candidate was successful in receiving a job offer from the company they have applied to. 

\emph{Adult Census Income:} The Adult Census Income dataset \cite{adult_dataset} is composed of demographic, educational and work-related information about 32,561 entries. Sensitive features include \textit{age}, \textit{race}, \textit{sex}, and more. The target indicates if individuals belonging to a certain entry in the dataset earn more than 50,000 US dollars per annum. Note that preprocessing was applied to this dataset whereby duplicates were removed and fields with missing values were replaced with \textit{unknown} for better interpretability.

\emph{Bank Marketing:} The Bank Marketing dataset \cite{bank_dataset} includes bank campaign-related and demographic information about 45,211 customers of a Portuguese banking institution. The sensitive feature for this data is \textit{age}. The binary classification task is to predict whether an individual subscribes to a term deposit or not.

The datasets are split into 67\% training and 33\% testing subsets. The training data is used to train a model that does not take fairness into account and biased predictions are generated on the testing subset. The resulting testing subset (now containing ground truth and biased predictions) is then fed to the pipeline which trains bias-mitigation models and generates bias-mitigated predictions.

\section{Results}
\label{results}

We focus on the quantitative evaluation measures related to performance and fairness along with qualitative results describing the explanations. We summarise accuracy and fairness measures of the base (biased) model, i.e., a random forest classifier, along with all the tested mitigation methods in Table \ref{table:metrics}.

\input{table_metrics}

\input{table_affected_cohorts_ratio}

\subsection{There is always a negatively impacted cohort}
 Table \ref{table:cohort_ratio} highlights that bias mitigation interventions always induce knock-on effects that negatively impact a cohort (labelled Disagree(-) Ratio). Outcomes for this group that were previously favourable have been changed to be unfavourable solely on account of mitigation interventions.
This result holds regardless of the bias-mitigation approach or dataset.
Additionally, on average, across all datasets and bias-mitigation approaches, the negatively impacted cohorts are larger than the positively impacted ones. This suggests that mitigation interventions have a high likelihood of harming individuals outside of the targeted group.

\subsection{Some fairness techniques do better than others in a consistent manner}

Our experiments suggest that some mitigation methods consistently perform better than others. 
Table \ref{table:metrics} shows that the Reject Option Classification (ROC) technique overall is better than the other methods tested. It demonstrates a good trade-off between accuracy and fairness measures (disparate impact in particular). On the other hand, Learning Fair Representations (LPR) consistently performs the worst in our battery of tests.
To quantify the extent of knock-on effects, Table \ref{table:cohort_ratio} displays the ratios of the people distributed in each of the affected cohorts. Here again, an ROC intervention impacts the smallest fraction of people, while LFR and Gerry Fair (GF) methods consistently induce the largest unintended effects.

A surprising observation for the Utrecht Fairness dataset in Table \ref{table:metrics} is that the in-processing techniques Gerry Fair (GF) and Prejudice Remover (PR) perform worse than the other models given that they usually offer the best trade-off between accuracy and fairness metrics due to the fact that they have access to more information while removing bias as compared to pre and post-processing techniques. This, however, is only observed for the Utrecht Fairness dataset and can be attributed to its small size.

\subsection{Explaining negatively affected groups}

We now discuss qualitative results stemming from our meta-classifier which identifies cohorts that are negatively impacted. 
Table \ref{table:cohort_ratio} shows relevant meta-classifier accuracy measures and class-specific precision. The lower performance of the model for the Utrecht dataset can be attributed to its smaller size. 
The precision of the decision tree for the affected cohorts follows the distribution of the cohort ratios as a result of the mapping after bias mitigation. That is, for large enough cohort ratios, the precision will be good as well. For example, the precision for the agreeable cohort for most datasets and bias-mitigation approaches is around 100\% and it follows the fact that the ratio of the agreeable cohort is composed of the majority of the population. Consequently, for the minority cohorts, the precision measurements can be quite low and sometimes even 0\%. Therefore, this method has the potential to work well if there is a meaningful sample for minority classes.

As an example, the following two decision rule sets were taken from the Utrecht dataset where Gerry Fair (GF) was used to provide bias-mitigated predictions that exhibit 33.9\% precision for the negatively affected cohort:

\textbf{(1)} \texttt{ind-languages <= 0.50 \& ind\_university\_grade <= 68.50 \& sport\_Rugby > 0.50 \& company\_C > 0.50 \& ind-degree\_bachelor > 0.50 \& ind-university\_grade > 55.00; [class: -1]}

\textbf{(2)} \texttt{ind-languages <= 0.50 \& ind-university\_grade > 68.50 \& gender\_female <= 0.50 \& company\_A <= 0.50 \& sport\_Swimming <= 0.50; [class: -1]}

The first decision rule set can be interpreted as follows: "Individuals who don't speak any foreign languages, have a bachelor's degree with a grade between 55 and 68, play Rugby and have applied for a job for company \textit{C} will receive a negative treatment.".

The second decision rule set can be interpreted as follows: "Male individuals who don't speak any foreign languages, who don't practice swimming, who have achieved a university grade > 68.5 and have not applied to company \textit{A} will receive a negative treatment.".

Furthermore, the following decision rule set is taken from the Adult dataset where Calibrated Equalized Odds (CEO) was used to provide bias-mitigated predictions that exhibit 17.8\% precision for the negatively impacted cohort:

\textbf{(1)} \texttt{Marital-status\_Married-civ-spouse <= 0.50 \& capital-gain <= 7139.50 \& hours-per-week > 43.50 \& age <= 43.50 \& (native-country\_Poland > 0.50 or native-country\_France > 0.50); [class: -1]}

This rule set can be interpreted as follows: "Individuals who are not married, are below 44 years of age, are of either Polish or French nationality, work above 43.5 hours per week and have an annual capital gain of less than 7140 currency units will receive a negative treatment.".

Thus, we can observe that we can identify the cohorts and the characteristics of individuals that will be impacted in a negative fashion with different success rates based on the ratio distribution among the different cohorts.

\section{Conclusion}
\label{conclusion}

In this paper, we looked at discovering and explaining the knock-on effects of bias mitigation by conducting an empirical study involving a number of biased datasets, various bias-mitigation approaches and a set of fairness metrics. We utilised two pipelines, one that performs bias mitigation and gives us bias-mitigated data and a bias-mitigator, and a subsequent one that utilises the bias-mitigated data in order to build an explainable classifier with which we could narrow down on the affected cohorts. Using the results of these pipelines, we showed three main things, (1) regardless of dataset and bias-mitigation approach utilised, there are always individuals who will receive a negative treatment post bias mitigation, (2) some bias-mitigation techniques perform better in comparison to others in a consistent fashion across all datasets by showing that ROC harms the least amount of individuals while also exhibits good accuracy and satisfies most fairness metrics, and (3) there are methods to allow for the discovery of the affected cohorts, one of which can be a decision tree classifier that will generate decision rule sets, the characteristics of which can describe the affected individuals and tell who will be impacted in future predicting scenarios. We also note that this method is reliant on the cohort ratio distributions and it performs best when there is a balance between the different classes. Thus, with this study, we argue that there is more to bias mitigation than only observing accuracy and fairness metrics as negative impacts can also manifest themselves in other shapes and forms and we stress on the more careful audit of static bias-mitigation interventions that go beyond taking into account only aggregate measurements.

\section{Acknowledgements}
This work was funded by the European Union’s Horizon Europe research and innovation programme under grant agreement no. 101070568 (AutoFair) and Science Foundation Ireland under Grant number 18/CRT/6183.


\medskip

{
\small
\bibliographystyle{plain}
\bibliography{mybib}
}

\end{document}

%% file: table_metrics.tex
\begin{table}[!ht]  
  \centering

\caption{Accuracy and fairness metrics across test sets for all datasets for biased and fairness models. Each metric result represents a tuple (x/y) where x is the mean while y is the standard deviation from a 5-fold cross-validation. (Note, the value in brackets for each fairness metric depicts its ideal value.)}
\label{table:metrics}


  \begin{tabular}{p{0.5cm} p{1.2cm} p{1.7cm} p{1.7cm} p{1.9cm} p{1.9cm} p{1.9cm} }
    \toprule
    & Model & Accuracy & Disparate Impact (1.0) & Average Odds (0.0) & Equal Opportunity (0.0) & Statistical Parity (0.0) \\ \midrule
    \parbox[t]{5mm}{\multirow{8}{*}{\rotatebox[origin=c]{90}{\textbf{Utrecht Fairness}}}} & Biased & 83.1\%/$\pm$1.7 & 0.75/$\pm$.49 & -0.13/$\pm$.19 & -0.27/$\pm$.33 & -0.08/$\pm$.13 \\
    & LFR   & 66.9\%/$\pm$7.1   & 1.43/$\pm$1.17   & 0.01/$\pm$.18  & -0.14/$\pm$0.15 & 0.08/$\pm$.22   \\
    & DIR   & 83.1\%/$\pm$3.1   & 0.63/$\pm$.36   & -0.13/$\pm$.19  & -0.17/$\pm$.39  & -0.12/$\pm$.13   \\
    & GF    & 74.0\%/$\pm$2.7   & 0.68/$\pm$.80   & -0.06/$\pm$.19  & -0.06/$\pm$.29 & -0.07/$\pm$.14   \\
    & PR    & 74.0\%/$\pm$1.4   & 1.28/$\pm$.83   & 0.05/$\pm$.20  & 0.10/$\pm$.31    & 0.02/$\pm$.15   \\
    & ROC   & 81.4\%/$\pm$3.8   & 0.80/$\pm$.23   & -0.08/$\pm$.16  & -0.09/$\pm$.32 & -0.09/$\pm$.10   \\
    & EO    & 76.8\%/$\pm$1.8   & 0.58/$\pm$.41   & -0.15/$\pm$.22  & -0.23/$\pm$.46 & -0.13/$\pm$.14   \\
    & CEO   & 77.0\%/$\pm$1.9   & 0.65/$\pm$.39   & -0.12/$\pm$.19  & -0.17/$\pm$.40 & -0.11/$\pm$.13   \\ \bottomrule
  \end{tabular}

\vspace{2pt}

  \begin{tabular}{p{0.5cm} p{1.2cm} p{1.7cm} p{1.7cm} p{1.9cm} p{1.9cm} p{1.9cm} }
    \toprule
           
    \parbox[t]{5mm}{\multirow{8}{*}{\rotatebox[origin=c]{90}{\textbf{Adult Census}}}} & Biased & 84.3\%/$\pm$1.0 & 0.55/$\pm$.12 & -0.05/$\pm$.06 & -0.04/$\pm$.12 & -0.11/$\pm$.03 \\
    & LFR   & 80.7\%/$\pm$1.3   & 0.57/$\pm$.12   & -0.06/$\pm$.04  & -0.08/$\pm$.09    & -0.08/$\pm$.03   \\
    & DIR   & 83.3\%/$\pm$1.3   & 0.55/$\pm$.07   & -0.04/$\pm$.03  & -0.01/$\pm$.07    & -0.10/$\pm$.02   \\
    & GF    & 83.0\%/$\pm$1.1   & 0.44/$\pm$.11   & -0.09/$\pm$.05  & -0.14/$\pm$.08    & -0.10/$\pm$.03   \\
    & PR    & 84.3\%/$\pm$1.1   & 0.39/$\pm$.06   & -0.13/$\pm$.03  & -0.20/$\pm$.05    & -0.13/$\pm$.02   \\
    & ROC   & 80.8\%/$\pm$2.0   & 1.13/$\pm$.26   & 0.11/$\pm$.07  & 0.14/$\pm$.08    & 0.04/$\pm$.07   \\
    & EO    & 79.5\%/$\pm$1.1   & 0.60/$\pm$.07   & -0.05/$\pm$.04  & -0.06/$\pm$.12    & -0.09/$\pm$.01   \\
    & CEO   & 79.5\%/$\pm$1.0   & 0.43/$\pm$.06   & -0.13/$\pm$.06  & -0.19/$\pm$.10    & -0.15/$\pm$.02   \\ \bottomrule
  \end{tabular}

\vspace{2pt}

  \begin{tabular}{p{0.5cm} p{1.2cm} p{1.7cm} p{1.7cm} p{1.9cm} p{1.9cm} p{1.9cm} }
    \toprule
           
    \parbox[t]{5mm}{\multirow{8}{*}{\rotatebox[origin=c]{90}{\textbf{Bank Marketing}}}} & Biased & 90.0\%/$\pm$0.0 & 1.36/$\pm$.30 & 0.0/$\pm$.03 & 0.0/$\pm$.06 & 0.02/$\pm$.01 \\
    & LFR   & 86.9\%/$\pm$2.4   & 1.05/$\pm$.45   & 0.0/$\pm$0.0  & 0.01/$\pm$.01    & 0.0/$\pm$0.0   \\
    & DIR   & 90.1\%/$\pm$0.0   & 1.22/$\pm$.16   & 0.0/$\pm$.02  & 0.01/$\pm$.03    & 0.01/$\pm$.01   \\
    & GF    & 90.0\%/$\pm$0.0   & 1.26/$\pm$.34   & 0.01/$\pm$.04  & 0.02/$\pm$.07    & 0.01$\pm$.01   \\
    & PR    & 90.2\%/$\pm$0.0   & 1.43/$\pm$.20   & 0.04/$\pm$.03  & 0.07/$\pm$.05    & 0.02$\pm$.01   \\
    & ROC   & 90.1\%/$\pm$0.0   & 1.24/$\pm$.14   & 0.0/$\pm$.04  & 0.0/$\pm$.07    & 0.02/$\pm$.01   \\
    & EO    & 90.0\%/$\pm$0.0   & 1.36/$\pm$.30   & 0.01/$\pm$.03  & 0.01/$\pm$.06    & 0.01/$\pm$.01   \\
    & CEO   & 90.0\%/$\pm$0.0   & 1.40/$\pm$.30   & 0.02/$\pm$.03  & 0.03/$\pm$.06    & 0.02/$\pm$.01   \\ \bottomrule
  \end{tabular}

\end{table}

%% file: table_affected_cohorts_ratio.tex
\begin{table}[!ht]  

\caption{Cohort ratio distribution post bias mitigation and precision for each cohort based on the meta classifier's predictions. Agreement shows the ratio of individuals whose treatment didn't change post bias mitigation. Positive disagreement (+) shows the ratio of individuals whose treatment changed from negative to positive while negative disagreement (-) shows the ratio of individuals whose treatment changed from positive to negative. The precisions for the cohorts show the percentage of predictions made by the meta classifier that are correct for that cohort. Each result in this table is a tuple (x/y) where x is the mean while y is the standard deviation from a 5-fold cross-validation.}

\scriptsize
\label{table:cohort_ratio}

  \centering

  \begin{tabular}{p{0.1cm} p{0.8cm} p{1.2cm} p{1.2cm} p{1.2cm} p{1.2cm} p{1.2cm} p{1.2cm}  p{1.2cm}}
    \toprule

           
    & Fairness Model & Agree \newline Ratio & Disagree(+) Ratio & Disagree(-) Ratio & Meta Clf \newline Accuracy  & Agree \newline Precision & Disagree(+) Precision & Disagree(-) Precision \\ \midrule

    \parbox[t]{5mm}{\multirow{7}{*}{\rotatebox[origin=c]{90}{\textbf{Utrecht Fairness}}}} & LFR & 66.9\%/$\pm$7.1 & 12.0\%/$\pm$4.7 & 21.1\%/$\pm$5.4 & 69.0\%/$\pm$6.0  & 96.6\%/$\pm$6.5 & 0.0\%/$\pm$0.0 & 14.2\%/$\pm$30.5 \\
    
    & DIR   & 83.1\%/$\pm$3.1     & 6.4\%/$\pm$1.3   & 10.4\%/$\pm$4.1  & 82.7\%/$\pm$2.5 & 99.4\%/$\pm$.06 & 0.0\%/$\pm$0.0 & 0.0\%/$\pm$0.0 \\
    & GF    & 74.0\%/$\pm$2.7   & 6.2\%/$\pm$2.1   & 19.8\%/$\pm$3.5     & 74.6\%/$\pm$4.1 & 90.0\%/$\pm$3.0 & 19.9\%/$\pm$13.6 & 33.9\%/$\pm$13.0 \\
    & PR    & 73.9\%/$\pm$1.4   & 7.4\%/$\pm$.08   & 18.7\%/$\pm$1.7     & 74.7\%/$\pm$3.6 & 92.8\%/$\pm$3.7 & 17.2\%/$\pm$14.6 & 26.3\%/$\pm$17.9 \\
    & ROC   & 81.5\%/$\pm$3.8   & 12.1\%/$\pm$1.3   & 6.4\%/$\pm$3.6      & 81.3\%/$\pm$3.8 & 99.4\%/$\pm$.06 & 2.2\%/$\pm$5.0 & 0.0\%/$\pm$0.0 \\
    & EO    & 76.8\%/$\pm$1.8   & 4.3\%/$\pm$.08   & 18.9\%/$\pm$2.4     & 77.2\%/$\pm$3.4 & 95.8\%/$\pm$1.4 & 0.0\%/$\pm$0.0 & 20.2\%/$\pm$13.9 \\
    & CEO   & 77.0\%/$\pm$1.9   & 4.3\%/$\pm$.07   & 18.7\%/$\pm$2.5       & 76.8\%/$\pm$3.6 & 95.0\%/$\pm$1.5 & 0.0\%/$\pm$0.0 & 20.7\%/$\pm$13.6 \\ \bottomrule
    
  \end{tabular}

\vspace{2pt}

  \begin{tabular}{p{0.1cm} p{0.8cm} p{1.2cm} p{1.2cm} p{1.2cm} p{1.2cm} p{1.2cm} p{1.2cm}  p{1.2cm}}
    \toprule
           
    \parbox[t]{5mm}{\multirow{7}{*}{\rotatebox[origin=c]{90}{\textbf{Adult Census}}}} & LFR & 80.7\%/$\pm$1.3 & 5.7\%/$\pm$.07 & 13.6\%/$\pm$1.2 & 80.7\%/$\pm$1.3 & 99.9\%/$\pm$0.0 & 0.0\%/$\pm$0.0 & 0.01\%/$\pm$.02 \\
    
    & DIR   & 83.2\%/$\pm$1.3   & 6.1\%/$\pm$.08   & 10.7\%/$\pm$.08 & 83.1\%/$\pm$1.4 & 99.8\%/$\pm$.02 & 0.0\%/$\pm$0.0 & 0.02\%/$\pm$.05 \\
    & GF    & 83.0\%/$\pm$1.1   & 4.6\%/$\pm$.07   & 12.4\%/$\pm$.05 & 84.0\%/$\pm$.07 & 99.1\%/$\pm$.04 & 3.4\%/$\pm$1.0 & 12.7\%/$\pm$2.6 \\
    & PR    & 84.3\%/$\pm$1.1   & 5.4\%/$\pm$.07   & 10.3\%/$\pm$.09 & 84.3\%/$\pm$1.1 & 99.2\%/$\pm$.02 & 12.4\%/$\pm$5.3 & 0.07\%/$\pm$.08 \\
    & ROC   & 80.8\%/$\pm$2.0   & 12.8\%/$\pm$1.9   & 6.4\%/$\pm$1.0 & 80.8\%/$\pm$1.9 & 99.9\%/$\pm$0.0 & 0.07\%/$\pm$.08 & 0.02\%/$\pm$.05 \\
    & EO    & 79.6\%/$\pm$1.0   & 3.7\%/$\pm$.04   & 16.7\%/$\pm$.08 & 80.0\%/$\pm$1.2 & 97.6\%/$\pm$1.1 & 0.0\%/$\pm$0.0 & 13.6\%/$\pm$5.4 \\
    & CEO   & 79.4\%/$\pm$1.0   & 3.7\%/$\pm$.07   & 16.9\%/$\pm$.05 & 80.2\%/$\pm$1.1 & 97.2\%/$\pm$.09 & 0.0\%/$\pm$0.0 & 17.8\%/$\pm$5.7 \\ \bottomrule
    
  \end{tabular}

\vspace{2pt}

 \begin{tabular}{p{0.1cm} p{0.8cm} p{1.2cm} p{1.2cm} p{1.2cm} p{1.2cm} p{1.2cm} p{1.2cm}  p{1.2cm}}
    \toprule

    \parbox[t]{5mm}{\multirow{7}{*}{\rotatebox[origin=c]{90}{\textbf{Bank Marketing}}}} & LFR & 86.9\%/$\pm$2.4 & 1.7\%/$\pm$2.9 & 11.4\%/$\pm$.05 & 87.9\%/$\pm$3.0 & 97.9\%/$\pm$1.1 & 0.0\%/$\pm$0.0 & 25.0\%/$\pm$1.3 \\
    
    & DIR   & 90.0\%/$\pm$.03   & 2.0\%/$\pm$.03  & 8.0\%/$\pm$.03 & 90.0\%/$\pm$.03 & 100\%/$\pm$0.0 & 0.0\%/$\pm$0.0 & 0.0\%/$\pm$0.0 \\
    & GF    & 90.0\%/$\pm$.04   & 1.8\%/$\pm$.03   & 8.2\%/$\pm$.04 & 89.4\%/$\pm$.04 & 97.8\%/$\pm$1.4 & 25.9\%/$\pm$11.0 & 11.6\%/$\pm$8.3 \\
    & PR    & 90.2\%/$\pm$.04   & 2.1\%/$\pm$.04   & 7.7\%/$\pm$.05 & 89.9\%/$\pm$.03 & 98.6\%/$\pm$.03 & 16.3\%/$\pm$7.9 & 6.8\%/$\pm$2.4 \\
    & ROC   & 90.1\%/$\pm$.03   & 4.2\%/$\pm$.06   & 5.7\%/$\pm$.04 & 90.1\%/$\pm$.03 & 100\%/$\pm$0.0 & 0.0\%/$\pm$0.0 & 0.0\%/$\pm$0.0 \\
    & EO    & 89.9\%/$\pm$.03   & 2.0\%/$\pm$.03   & 8.1\%/$\pm$.05 & 89.9\%/$\pm$.03 & 100\%/$\pm$0.0 & 0.0\%/$\pm$0.0 & 0.01\%/$\pm$.03 \\
    & CEO   & 90.0\%/$\pm$.02   & 2.2\%/$\pm$.02   & 7.8\%/$\pm$.03 & 90.0\%/$\pm$.02 & 99.9\%/$\pm$0.0 & 0.0\%/$\pm$0.0 & 0.02\%/$\pm$.03 \\ \bottomrule
    
  \end{tabular}

\end{table}